\documentclass[11pt,a4paper]{article}
\usepackage{authblk}
\usepackage[hyphens]{url}
\usepackage[hyperref]{acl2020}
\usepackage{times}
\usepackage{latexsym}
\usepackage{amsmath}
\usepackage{amssymb}
\usepackage{tabularx}
\usepackage{graphicx}
\usepackage{mathtools}

\usepackage{multirow}

\usepackage{microtype}

\urldef{\footurl}\url{https://github.com/AndrewJGaut/Towards-Understanding-Gender-Bias-in-Neural-Relation-Extraction}

\definecolor{xxx}{RGB}{0,127,255}

\makeatletter
\newcommand\email[2][]%
   {\newaffiltrue\let\AB@blk@and\AB@pand
      \if\relax#1\relax\def\AB@note{\AB@thenote}\else\def\AB@note{\relax}%
        \setcounter{Maxaffil}{0}\fi
      \begingroup
        \let\protect\@unexpandable@protect
        \def\thanks{\protect\thanks}\def\footnote{\protect\footnote}%
        \@temptokena=\expandafter{\AB@authors}%
        {\def\\{\protect\\\protect\Affilfont}\xdef\AB@temp{#2}}%
         \xdef\AB@authors{\the\@temptokena\AB@las\AB@au@str
         \protect\\[\affilsep]\protect\Affilfont\AB@temp}%
         \gdef\AB@las{}\gdef\AB@au@str{}%
        {\def\\{, \ignorespaces}\xdef\AB@temp{#2}}%
        \@temptokena=\expandafter{\AB@affillist}%
        \xdef\AB@affillist{\the\@temptokena \AB@affilsep
          \AB@affilnote{}\protect\Affilfont\AB@temp}%
      \endgroup
       \let\AB@affilsep\AB@affilsepx
}
\makeatother

\newcommand\blfootnote[1]{%
  \begingroup
  \renewcommand\thefootnote{}\footnote{#1}%
  \addtocounter{footnote}{-1}%
  \endgroup
}

\aclfinalcopy 

\newcolumntype{x}[1]{%
>{\raggedright\hspace{0pt}}p{#1}}%
\newcolumntype{P}[1]{>{\centering\arraybackslash}p{#1}}
\newcolumntype{M}[1]{>{\centering\arraybackslash}m{#1}}

\newcommand{\tnhl}{\tabularnewline\hline}

\title{Towards Understanding Gender Bias in Neural Relation Extraction}

\author[$\dagger$]{\textbf{Andrew Gaut}*}
 \author[$\dagger$]{\textbf{Tony Sun}*}
\author[$\dagger$]{\textbf{Shirlyn Tang}}
\author[$\dagger$]{\textbf{Yuxin Huang}}
\author[$\dagger$]{\\\textbf{Jing Qian}}
\author[$\dagger \dagger$]{\textbf{Mai ElSherief}}
\author[$\ddagger$]{\textbf{Jieyu Zhao}}
\author[$\dagger$]{\\\textbf{Diba Mirza}}
\author[$\dagger$]{\textbf{Elizabeth Belding}}
\author[$\ddagger$]{\textbf{Kai-Wei Chang}}
\author[$\dagger$]{\textbf{William Yang Wang}}
\affil[$\dagger$]{Department of Computer Science, UC Santa Barbara}
\affil[$\ddagger$]{Department of Computer Science, UC Los Angeles}
\affil[$\dagger \dagger$]{School of Interactive Computing, Georgia Institute of Technology}

\email{ \texttt{\{ajg, tonysun, shirlyntang, yuxinhuang\}@ucsb.edu} }
\email{  \texttt{\{jing\_qian, dimirza, ebelding, william\}@cs.ucsb.edu}}
\email{  \texttt{melsherief@gatech.edu}}
\email{ \texttt{\{jyzhao, kwchang\}@cs.ucla.edu} }

\date{}

\begin{document}
\maketitle
\begin{abstract}

    Recent developments in Neural Relation Extraction (NRE) have made significant strides towards automated knowledge base construction. While much attention has been dedicated towards improvements in accuracy, there have been no attempts in the literature to evaluate social biases exhibited in NRE systems. In this paper, we create WikiGenderBias, a distantly supervised dataset composed of over 45,000 sentences including a 10\% human annotated test set for the purpose of analyzing gender bias in relation extraction systems. We find that when extracting spouse and hypernym (i.e., occupation) relations, an NRE system performs differently when the gender of the target entity is different. However, such disparity does not appear when extracting relations such as birth date or birth place. We also analyze two existing bias mitigation techniques, word embedding debiasing and data augmentation. Unfortunately, due to NRE models relying heavily on surface level cues, we find that existing bias mitigation approaches have a negative effect on NRE. Our analysis lays groundwork for future quantifying and mitigating bias in relation extraction.






\end{abstract}

\section{Introduction}
\label{intro}

\blfootnote{* Equal Contribution.}

With the wealth of information being posted online daily, relation extraction has become increasingly important. Relation extraction aims specifically to extract relations from raw sentences and represent them as succinct relation tuples of the form \textit{(head, relation, tail)} e.g.,  \textit{(Barack Obama, spouse, Michelle Obama)}. 

The concise representations provided by relation extraction models have been used to extend Knowledge Bases (KBs) \cite{riedel2013relation, subasic2019building, trisedya2019neural}. These KBs are then used heavily in NLP systems, such as question answering systems \cite{bordes2014question, yin2015neural, cui2019kbqa}. In recent years, much focus in the Neural Relation Extraction (NRE) community has been centered on improvements in model precision and the reduction of noise \cite{lin2016neural,liu2017soft, wu2017adversarial,feng2018reinforcement, vashishth2018reside,qin2018robust}. Yet, little attention has been devoted towards the fairness of such systems.

\begin{figure*}
\centering
{\includegraphics[width=0.9\textwidth, height = 4.5cm]{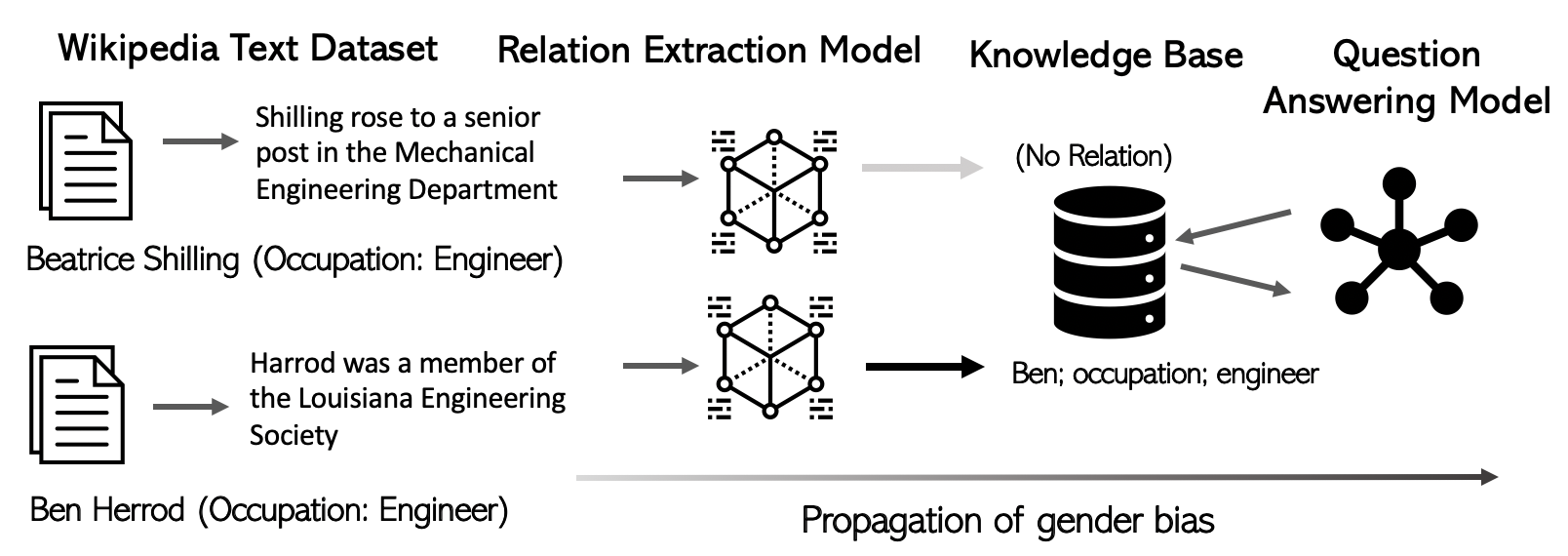}}
\caption{An illustration of gender bias in relation extraction and how it affects a downstream application. In their Wikipedia articles, both Beatrice (female) and Ben (male) are described as engineers. These sentences contain the \textit{(entity; occupation; engineer)} relation. However, the model only predicts that the sentence from the male article expresses the occupation relation. If on a large scale, models extract the \textit{(entity; occupation; engineer)} relation more often for males, knowledge bases will contain information for male engineers more often than female. Question answering models that query these knowledge bases may give biased answers and propagate gender bias downstream.}
\label{relation_extraction_bias}
\end{figure*}

We take the first step at understanding and evaluating gender bias in NRE systems by measuring the differences in model performance when extracting relations from sentences written about females versus sentences written about males. If a NRE model predicts a relation such \textit{occupation} with higher recall on male entities, this could lead to the resulted knowledge bases having more occupation information for males than for females (see the illustration in Figure \ref{relation_extraction_bias}). 
Eventually, the gender bias in knowledge bases may affect downstream predictions, causing undesired allocative harms \cite{crawford2018trouble} and
reinforcing gender-stereotypical beliefs in society.

In this paper, we present an evaluation framework to analyze social bias in NRE models. Specifically, we evaluate gender bias in English language predictions of a collection of popularly used and open source NRE models\footnote{\footurl} \cite{lin2016neural, wu2017adversarial, liu2017soft, feng2018reinforcement}. We evaluate on two fronts: (1) examining gender bias exhibited in a model that is trained on a relation extraction dataset; 
and (2) examining if the existing bias mitigation techniques~\cite{bolukbasi2016man, zhao2018gender, lu2018gender} can be applied to reduce the bias in an NRE system while maintaining its performance. 

Carrying out such an evaluation is difficult with existing NRE datasets, such as the NYT dataset~\cite{sandhaus2018nyt}, because there is no reliable way to obtain gender information about the entities mentioned in input sentences. Therefore, we create a new dataset, WikiGenderBias, specifically aimed at evaluating gender bias for NRE \footnote{The dataset can be found here: \url{https://github.com/AndrewJGaut/Towards-Understanding-Gender-Bias-in-Neural-Relation-Extraction}}.  
WikiGenderBias is a distantly supervised dataset extracted using Wikipedia and DBPedia. It contains 45,000 sentences, each of which describe either a male or female entity with one of four relations: {\it spouse, hypernym (i.e., occupation), birthDate, and birthPlace}. 
We posit that a biased NRE system leverages gender information as a proxy when extracting knowledge tuples with spouse and hypernym relations. However, gender of the entity does not affect the extraction of relations such as birthDate and birthPlace, as they are not intuitively related to gender. Experiment results confirm our conjecture. 


Our contributions are as such:

\begin{itemize}
    \item We create WikiGenderBias, a new dataset for evaluating gender bias in NRE systems.
    \item We present an evaluation framework to demonstrate that gender bias is exhibited in NRE model outputs.
    \item We test several existing bias mitigation approaches to reducing gender bias in NRE system. Our analysis sheds light for designing future mitigating techniques.
    
    
    
    
\end{itemize}

\begin{table*}
\centering
  \begin{tabular}{|l|c c|c c|c c|c c|c c|c c|}
    \hline
    \multirow{2 }{*} &
      \multicolumn{4}{c|}{Original Dataset} &
      \multicolumn{4}{c|}{Equalized Dataset}\\
    \hline
    \multirow{2}{*} &
      \multicolumn{2}{c|}{Entity Pairs} &
      \multicolumn{2}{c|}{Instances} &
      \multicolumn{2}{c|}{Entity Pairs} &
      \multicolumn{2}{c|}{Instances}\\
    & M & F & M & F & M & F & M & F \\
    \hline
Train & 12,139 & 4,571 & 27,048 & 9,391& 2,479 & 4,571 & 9,465 & 9,415\\ 
 \hline 
Development & 1,587 & 553 & 3,416 & 1,144& 336 & 553 & 1,144 & 1,144\\ 
 \hline 
Test & 1,030 & 1,101 & 2,320 & 2,284 & 1,030 & 1,101 & 2,320 & 2,284\\ 
 \hline 
 Total & 14,756 & 6,225 & 32,784 & 12,819 & 3,845 & 6,225 & 12,929 & 12,843 \\
 \hline
  \end{tabular}
  \caption{WikiGenderBias's Dataset splits. Entity Pairs means distinct pairs $(e1,e2)$ such that $(e_1, relation, e_2)$ is a relation in WikiGenderBias. Instances are the total number of $(e1, relation, e2, sentence)$ tuples in WikiGenderBias, where $sentence$ is distantly supervised. We categorize an entity pair as male (female) if $e1$ is male (female), since the sentence in the instance is taken from $e1$'s article and we define datapoints as male (female) if that is the gender of the subject of the article. The left two entries are for the dataset taken from the true distribution; the right two are the gender-equalized dataset created by down-sampling male instances.}
\label{WikiGenderBias}
\end{table*}

\section{Related Work}
\label{related_work}

\paragraph{Gender Bias Measurement.} Existing studies have revealed gender bias in various NLP tasks \cite{zhao2017men, rudinger2018gender, zhao2018gender, dixon2018measuring, lu2018gender, kiritchenko2018examining, romanov2019s, sheng-etal-2019-woman, sun2019mitigating}. 
People have proposed different metrics to evaluate gender bias, for example,
by using the performance difference of the model on male and female datapoints for bias evaluation~\cite{lu2018gender, kiritchenko2018examining}. 
Other metrics have been proposed to evaluate fairness of predictors and allocative bias \cite{dwork2012fairness, hardt2016equality}, such as Equality of Opportunity. In this work, we use both of these metrics to evaluate NRE models.

\paragraph{Mitigation Methods.} After discovering gender bias existing, prior work has developed various methods to mitigate that bias \cite{escude-font-costa-jussa-2019-equalizing, bordia-bowman-2019-identifying}. Those mitigation methods can be applied in different levels of a model, including in the training phase, in the embedding layer, or in the inference procedure. In this paper, we test three existing debiasing approaches, namely data augmentation \cite{zhao2018gender, lu2018gender}, and word embedding debiasing technique (Hard Debiasing \cite{bolukbasi2016man}) for mitigating bias in NRE models.

\paragraph{Neural Relation Extraction.} Relation extraction is a task in NLP with a long history that typically seeks to extract structured tuples $(e_1, r, e_2)$ from texts \cite{bach2007review}. Early on, learning algorithms for relation extraction models were typically categorized as supervised, including feature-based methods \cite{kambhatla2004combining, guodong2005exploring, zhao2005extracting} and kernel-based methods \cite{lodhi2002text, zelenko2003kernel}, or semi-supervised~\cite{brin1998extracting, agichtein2000snowball, etzioni2005unsupervised, pantel2006espresso}, or purely unsupervised  \cite{etzioni2008open}. Supervised approaches suffer from the need for large amounts of labelled data, which is sometimes not feasible, and generalizes poorly to open domain relation extraction, since labeled data is required for every entity-relation type \cite{bach2007review, mintz2009distant}. Many semi-supervised approaches rely on pattern-matching, which is not robust, and many are unable to extract intra-sentence relations \cite{bach2007review}. When data annotation is insufficient or hard to obtain and semi-supervised approaches are insufficient, the distant supervision assumption is used to collect data to train supervised models \cite{mintz2009distant}. Given a relation $(e_1, r, e_2)$ in a knowledge base (KB), distant supervision assumes any sentence that contains both $e_1$ and $e_2$ expresses $r$ \cite{mintz2009distant}. 
Great efforts have been made to improve NRE models by mitigating the effects of noise in the training data introduced by Distant Supervision~\cite{hoffmann2011knowledge, surdeanu2012multi, lin2016neural, liu2017soft, feng2018reinforcement, qin2018robust}. However, to our knowledge, there are no studies on bias or ethics in NRE, which is filled by this work.

\section{WikiGenderBias}

We define gender bias in NRE as a difference in model performance when predicting on sentences from male versus female articles. Thus, we need articles written about entities for which we can identify the gender information. However, to obtain gender information for existing annotated datasets could be costly or impossible. Thus, we elected to create WikiGenderBias with this gender information to be able to detect scenarios like that in Figure \ref{relation_extraction_bias}. The data statistics of WikiGenderBias are given in Table \ref{WikiGenderBias}.

\subsection{Dataset Creation}
\label{section_dataset_creation}

Wikipedia is associated with a knowledge base, DBPedia, that contains relation information for entities with articles on Wikipedia \cite{lrec12mendes2}. Many of these entities have gender information and their corresponding articles are readily available. Therefore, we create our dataset based on sentences extracted from Wikipedia.

To generate WikiGenderBias, we use a variant of the distant supervision assumption: for a given relation between two entities, if one sentence from an article written about one entity also mentions the other entity, then we assume that such sentence expresses the relation. For instance, if we know \textit{(Barack, spouse, Michelle)} is a relation tuple and we find the sentence \textit{He and Michelle were married} in Barack's Wikipedia article, then we assume that sentence expresses the \textit{(Barack, spouse, Michelle)} relation. This assumption is similar to that made by \citet{mintz2009distant} and allows us to scalably create the dataset.

\begin{table*}[t]
\centering
\small
\footnotesize
\def\arraystretch{1.8}
\begin{tabular}{|x{1.5cm}|x{2cm}|x{2cm}|x{8cm}|}
    \hline
    \textbf{Relation} & \textbf{Head Entity} & \textbf{Tail Entity} & \textbf{Sentence}
	\tnhl
    Birthdate  & Robert M. Kimmitt  &    December 19, 1947   &     Robert M. Kimmitt ( born December 19 , 1947 ) was United States Deputy Secretary of the Treasury under President George W. Bush . 
\tnhl
Birthplace &    Charles Edward Stuart & Rome  & Charles was born in the Palazzo Muti , Rome , Italy , on 31 December 1720 ,  where his father had been given a residence by Pope Clement XI 
\tnhl
Spouse    &   John W. Caldwell      &     Sallie J. Barclay    &          Caldwell married Sallie J. Barclay , and the couple had one son and two daughters . 
\tnhl

 hypernym  & Handry Satriago     &      CEO                  &                       Handry Satriago ( born in Riau , Pekanbaru on June 13 , 1969 ) is the CEO of General Electric Indonesia . 
\tnhl
\end{tabular}
\caption{Examples of relations of each type in WikiGenderBias.}
\label{tasks_biases}
\end{table*}

\begin{figure}
\centering
{\includegraphics[width=0.5\textwidth, height = 4cm]{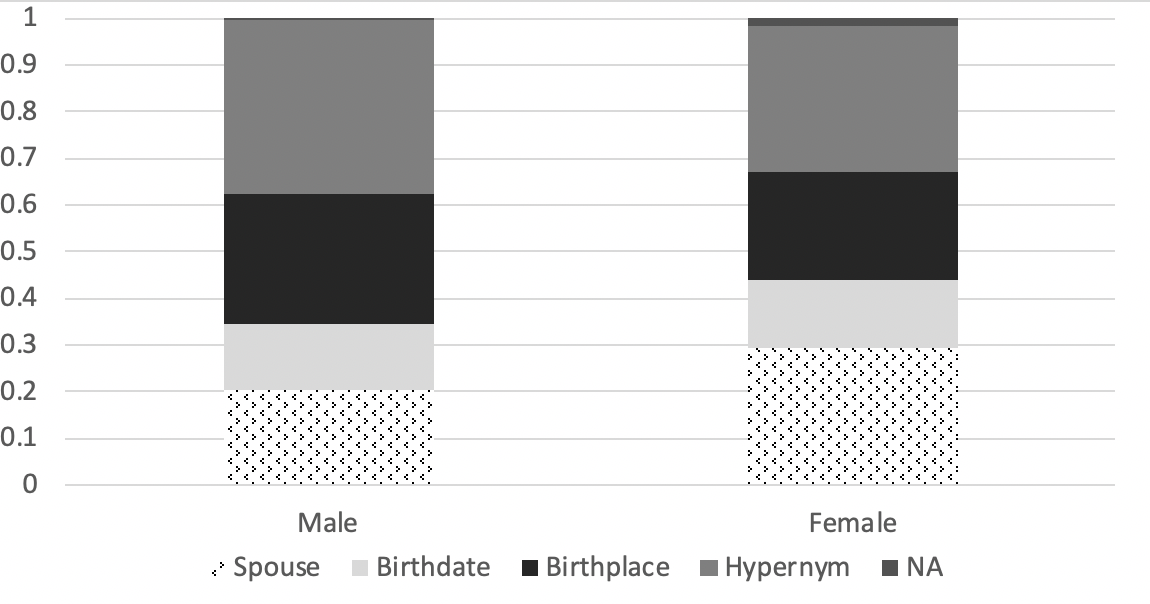}}
\caption{Proportion of sentences corresponding to a given relation over total sentences in WikiGenderBias for each entity. This demonstrates that, of the entities we sampled to create WikiGenderBias, the spouse relation is expressed more often relative to the birthdate, birthplace, and hypernym relations in articles about female entities than in articles about male entities. Additionally, hypernym is mentioned more often relative to the other relations in articles about male entities than in articles about female entities.}
\label{gender_proportion}
\end{figure}

WikiGenderBias considers four relations that stored in DBPedia: \textit{spouse}, \textit{hypernym}, \textit{birthDate}, and \textit{birthPlace}. 
Note that the hypernym relation on DBPedia is similar to occupation, with entities having hypernym labels such as Politican. We also generate negative examples by obtaining datapoints for three unrelated relations: \textit{parents}, \textit{deathDate}, and \textit{almaMater}. We label them as NA (not a relation). As each sentence only labelled with one relation based on our distant supervision assumption, WikiGenderBias is a 5-class classification relation extraction task. Figure \ref{gender_proportion} lists the label distribution. 

We hypothesize that a biased relation extraction model might use gender as a proxy to influence predictions for spouse and hypernym relations, since words pertaining to marriage are more often mentioned in Wikipedia articles about female entities and words pertaining to hypernym (which is similar to occupation) are more often mentioned in Wikipedia articles about male entities \cite{wagner2015s, graells2015first}. On the other hand, we posit that birthDate and birthPlace would operate like control groups and believe gender would correlate with neither relation. 

To simplify the analysis, we only consider the head entities that associated with at least one of the four targeted relations. 
We set up our experiment such that head entities are not repeated across the train, dev, and test sets so that the model will see only new head entities at the test time. Since we obtain the distantly supervised sentences for a relation from the head entity's article, this guarantees the model will not reuse sentences from an article. However, it is possible that the head entity will appear as a tail entity in other relations because an entity could appear in multiple articles. The data splits are given in Table \ref{WikiGenderBias}. 

Besides, Wikipedia includes more articles written about males than about females. Therefore, there are more male instances than female instances in WikiGenderBias as well. To remove the effect of dataset bias in our analysis, we also create a gender-equalized version of the training and development sets by down-sampling male instances.
We discuss the creation of gender-equalized test set below.

\subsection{Test Sets}
\label{wiki_test_sets}

We equalize the male and female instances in the test set. In this way, a model cannot achieve high performance by performing well only on the dominant class. 
Furthermore, since some data instances that are collected using distant supervision are noisy, 
 we annotated the correctness of the test instances using Amazon Mechanical Turk annotations to perform a fair comparison. 

Specifically, we asked workers to determine whether or not a given sentence expressed a given relation. If the majority answer was ``no'', then we labeled that sentence as expressing ``no relation'' (we denote them as NA). Each sentence was annotated by three workers. Each worker was paid 15 cents per annotation. We only accepted workers from England, the US or Australia and with HIIT Approval Rate greater than $95\%$ and Number of HIITs greater than $100$. We found the pairwise inter-annotator agreement as measured by Fleiss' Kappa \cite{fleiss1971measuring} $\kappa$ is 0.44, which is consistent across both genders and signals moderate agreement. We note that our $\kappa$ value is affected by asking workers to make binary classifications, which limits the degree of agreement that is attainable above chance. We also found the pairwise inter-annotator agreement to be 84\%.

\subsection{Data Analysis}
\label{WikiGenderBias_results}

We build on the work of \citet{graells2015first}, who discovered that female entities are more likely to have spouse information in the Infoboxes on their Wikipedia page than male entities. Figure \ref{gender_proportion} demonstrates a further discrepancy: amongst articles we sampled, proportionally, the spouse relation is mentioned more often relative to hypernym, birthPlace, and birthDate in female articles than in male articles. Additionally, we show that amongst female and male articles we sampled, hypernyms are mentioned more often in male than female articles relative to spouse, birthPlace, and birthDate (see Section \ref{gender_proportion}). This observation aligns with the literature, arguing that 
authors do not write about the two genders equally \cite{wagner2015s, graells2015first}.

\section{Gender Bias in NRE}

We evaluate OpenNRE \cite{han2019opennre}, a popular open-source NRE system. OpenNRE implements the approach from \cite{lin2016neural}. To convert sentences into vectors, researchers propose convolutional neural networks  as well as the pieceweise convoultional neural networks (PCNN) which retain more structural information between entities \cite{zeng2015distant}. In this work, we use a PCNN with Selective Attention for the experiments.

We train every encoder-selector combination on the training set of WikiGenderBias and its gender-equalized version. We input Word2Vec \cite{mikolov2013distributed} word embeddings trained on WikiGenderBias to the models\footnote{We performed Grid Search to determine the optimal hyperparameters. We set epochs$=60$, learning rate $\eta=0.5$, early stopping with patience of 10, \textit{batch size}$=160$, and \textit{sliding window size}$=3$ (for CNN and PCNN). These hyperparameters are similar to the default settings found in the OpenNRE repository tensorflow branch, which uses epochs$=60$, learning rate $\eta=0.5$, and early stopping with patience of 20.}. We use commit 709b2f from the OpenNRE repository tensorflow branch to obtain the models. 
\begin{figure*}
\centering
  \includegraphics[width=0.75\textwidth, height=4cm]{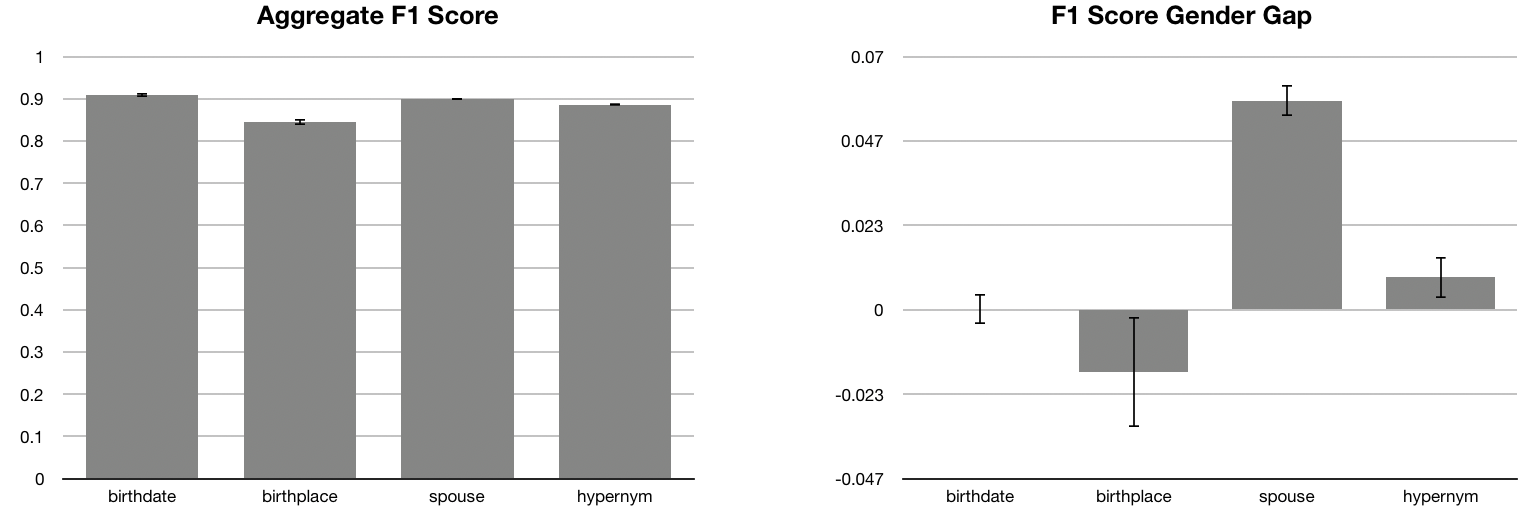}
  \caption{Aggregate performance of the NRE model for each relation (left) and $male - female$ F1 score gender gap for each relation (right). An ideal model maximizes performance and minimizes the gender gap. The experiment is run five times. We give the mean values and standard error bars.}
  \label{no_debiasing_performance}
\end{figure*}

\subsection{Performance Parity Score}
\label{harmonic_metric}
The goal of a successful relation extraction model is to maximize F1 score while minimizing the model performance gender gap (or disparity score). However, when comparing different systems, it is hard to decide what is the right balance between these two objectives. 
On one end, a model which has zero gender gap but has only 10\% accuracy for both male and female test instances has almost no practical value. Other methods that have high accuracy or F1 score may do so at the cost of a wide gender gap. Although our test set for WikiGenderBias is gender-equalized, one can imagine that improving performance on a test set that is heavily skewed towards males can be done by focusing on male test instances while largely ignoring female ones. Therefore, it is important to strike a balance between model performance and inter-group parity.

To measure model performance, we use Macro-average F1 score. To measure inter-group parity, we use the pairwise difference in F1 scores averaged over all the groups for predictions on a given relation $i$. We describe the average difference over all relations as Disparity Score (DS):
\begin{align*}
    DS = \frac{1}{n}  \sum_{i=1}^{n} \frac{1}{x}  \sum_{j=1}^{x} \sum_{k=j+1}^{x}\biggl|F_{1_{ik}} - F_{1_{ij}}\biggr|, 
\end{align*}
where $n$ denotes the number of relations (e.g. \{birthDate, birthPlace, spouse, hypernym\}). $x$ denotes the number of groups (e.g. \{male, female\}). $F_{1_{rk}}$ is the $F_{1}$ score for the model when predicting datapoints with true label relation $r$ that belong to group $k$. (So, for instance, $F_{1_{spouse,male}}$ is the $F_1$ score on sentences that express the spouse relation from male articles.) The Disparity Score measures the F1 score gap between predictions on male and female data points.

Bringing these two metrics together, we propose the Performance Parity Score (PPS). PPS is the Macro-average difference (equally weighted) of the F1 score subtracted by the model performance gender gap, which we defined as the Disparity Score, per relation.  We place equal importance on the F1 score and Disparity Score by giving each score an implicit weight of $1$. In our formula for PPS above, we also divide the final result by the number of relations $n$. This keeps the range of PPS within $(-1,1]$, although PPS will generally fall between $[0,1]$ because it is highly unlikely that the Disparity Score will be greater than the overall F1 score. PPS seeks to incentivize a combination of both model performance and inter-group parity for the task of relation extraction:

{\small
\begin{align*}
    PPS & = {} \frac{1}{n} \sum_{i=1}^{n} \bigg(F_{1_i} - \frac{1}{x} \sum_{j=1}^{x} \sum_{k=j+1}^{x}\biggl|F_{1_{ik}} - F_{1_{ij}}\biggr| \bigg) \\
    &= \frac{1}{n} \sum_{i=1}^{n} F_{1_i} \! - \! \frac{1}{n}  \sum_{i=1}^{n} \frac{1}{x}  \sum_{j=1}^{x} \sum_{k=j+1}^{x} \biggl| F_{1_{ik}} \!-\! F_{1_{ij}}\biggr|  \\
    &= \mbox{Macro $F_1$ score} - \mbox{Disparity Score}.
\end{align*}
}


\subsection{Measuring Performance Differences}
Similar to the parity term in PPS, gender bias can be measured as the difference in a performance metric for a model when evaluated on male and female datapoints \cite{de2019bias}. We define male (female) datapoints to be relations for which the head entity is male (female), which means the distantly supervised sentence is taken from a male (female) article. Prior work has used area under the precision-recall curve and F1 score to measure NRE model performance \cite{gupta2019neural, han2019opennre, kuang2019improving}. We use Macro-F1 score as our performance metric. We denote the F1 gender difference as $F1_{Gap}$, which is used to calculate the disparity score. A larger disparity score indicates higher bias in predictions.

\subsection{Equality of Opportunity Evaluation}
\label{eq_of_opp_subsec}

Equality of Opportunity (EoO) was originally proposed to measure and address allocative biases \cite{hardt2016equality}. Consequently, we examine this metric in the context of relation extraction to better understand how allocative biases can begin to emerge at this stage.

Equality of Opportunity (EoO) is defined in terms of the joint distribution of $(X, A, Y)$, where $X$ is the input, $A$ is a protected attribute that should not influence the prediction, and $Y$ is the true label \cite{hardt2016equality}. A predictor satisfies Equality of Opportunity if and only if: 
$P(\hat{Y}=1 | A=male, Y=1)$
and 
$P(\hat{Y}=1 | A=female, Y=1)$. In our case $A = \{male,female\}$, because gender is our protected attribute and we assume it to be binary. We evaluate EoO on a per-relation, one-versus-rest basis. Thus, when calculating EoO for spouse, $Y=1$ indicates the true label is spouse and $\hat{Y}=1$ indicates a prediction of spouse. We do this for each relation. Note that this is equivalent to measuring per-relation recall for each gender.

\begin{table*}[!t]
\centering
\small
 \begin{tabular}{@{}r@{ }  c c c  c c c  c c c  c c c c c c c@{}}
 \hline
 & \multicolumn{2}{c}{Spouse} &  \multicolumn{2}{c}{Birth Date} & \multicolumn{2}{c}{Birth Place} &  \multicolumn{2}{c}{Hypernym} &  \multicolumn{3}{c}{Total} \\\
& $F1_{Gap}$ & EoO & $F1_{Gap}$ & EoO & $F1_{Gap}$ & EoO & $F1_{Gap}$ & EoO & F1 Score & Disparity Score & PPS  \\
\hline
PCNN,ATT & .041 & .058& .004 & .000& -.003 & -.017& .015 & .009& .886 & .016 & .870\\ 
CNN,ATT & .034 & .043& -.003 & .001& .014 & .004& .028 & .014& .882 & .020 & .862\\ 
RNN,ATT & .032 & .043& .015 & .019& .005 & -.011& -.006 & -.006& .889 & .014 & .875\\ 
BIRNN,ATT & .039 & .061& .013 & .021& -.016 & -.033& -.013 & -.026& .884 & .020 & .864\\ 
\hline 
PCNN,AVE & .034 & .044& .005 & .010& -.001 & -.011& .005 & -.005& .903 & .011 & .892\\ 
CNN,AVE & .027 & .028& .013 & .029& .007 & .009& .002 & -.028& .895 & .012 & .883\\ 
RNN,AVE & .039 & .036& .004 & .021& .016 & .020& .006 & -.012& .912 & .016 & .895\\ 
BIRNN,AVE & .024 & .018& .001 & .015& .009 & .018& -.005 & -.022& .913 & .010 & .903\\ 
\hline \end{tabular}

\caption{Results from running combinations of encoders and selectors of the OpenNRE model for the male and female genders of each relation. A positive $F1_{Gap}$ indicates a higher F1 on male instances. A higher Equality of Opportunity (EoO) indicates higher recall on male instances. A higher PPS score indicates a better balance of performance and parity (see Section \ref{harmonic_metric}). We ran the experiment five times and report the mean values. Varying the encoder and selector appears to have no conclusive effect on bias, although models using the average selector doe achieve better aggregate performance. These results were obtained using the gender unequalized training data. 
}
\label{enc_sel_combos_truedistrib}
\end{table*}

\subsection{Result}

As shown in Figure \ref{no_debiasing_performance}, the NRE system performs better when predicting the spouse relation on sentences from articles about male entities than from articles on female entities (see Figure 3, right). Further, there is a large recall gap (see EoO column, row 1, in Table \ref{enc_sel_combos_truedistrib}). Notably, the gender difference in performance is much smaller on birthDate, birthPlace, and hypernym relations, although the gender difference is non-zero for birthPlace and hyerpym. This is interesting given that a higher percentage of female instances in WikiGenderBias are spouse relations than male (see Figure \ref{gender_proportion}). We encourage future work to explore whether the writing style differences between male and female spouse instances causes those male instances to be easier to classify.

 In addition, we explore different types of sentence encoder and sentence-level attention used in the creation of the bag representation for each entity pair and examined how these models performed on our dataset. Notably, the bias in spouse relation persists across OpenNRE architectures (see Table \ref{enc_sel_combos_truedistrib}). It seems models using average attention, which merely averages all the sentence vectors in the bag to create a representation of the entire bag, allows for better aggregate performance on WikiGenderBias. However, the effect on the Disparity Score (and therefore the bias exhibited in the predictions) seems negligible.
 
 We note that these results do not necessarily indicate that the model itself contains biases given that males and females are written about differently on Wikipedia. These results do, however, demonstrate that we must be cautious when deploying NRE models, especially those trained on Wikipedia data, since they can propagate biases latent in their training data to the knowledge bases they help create.

\section{Bias Mitigation}

We examine data augmentation and Hard-Debiasing as bias mitigation techniques for reducing gender bias in NRE system.

\subsection{Bias Mitigation Techniques}

\paragraph{Equalizing the Gender Distribution}

Sometimes, the true distribution contains an imbalance in gendered data. For instance, perhaps the training set contains more instances from male articles than female. To mitigate this, one can simply downsample the male instances until the male and female instances are approximately equal, then train on this modified, equalized distribution.

\paragraph{Data Augmentation.} The contexts in which males and females are written about can differ; for instance, on Wikipedia women are more often written about with words related to sexuality than men \cite{graells2015first}. Data augmentation mitigates these contextual biases by replacing masculine words in a sentence with their corresponding feminine words and vice versa for all sentences in a corpus, and then training on the union of the original and augmented corpora\footnote{We use the following list to perform data augmentation: \url{https://github.com/uclanlp/corefBias/blob/master/WinoBias/wino/generalized_swaps.txt} } \cite{zhao2018gender, lu2018gender, dixon2018measuring, maudslay2019s, zhao2019gender}.

\begin{figure*}
\centering
  \includegraphics[width=0.75\textwidth, height=6cm]{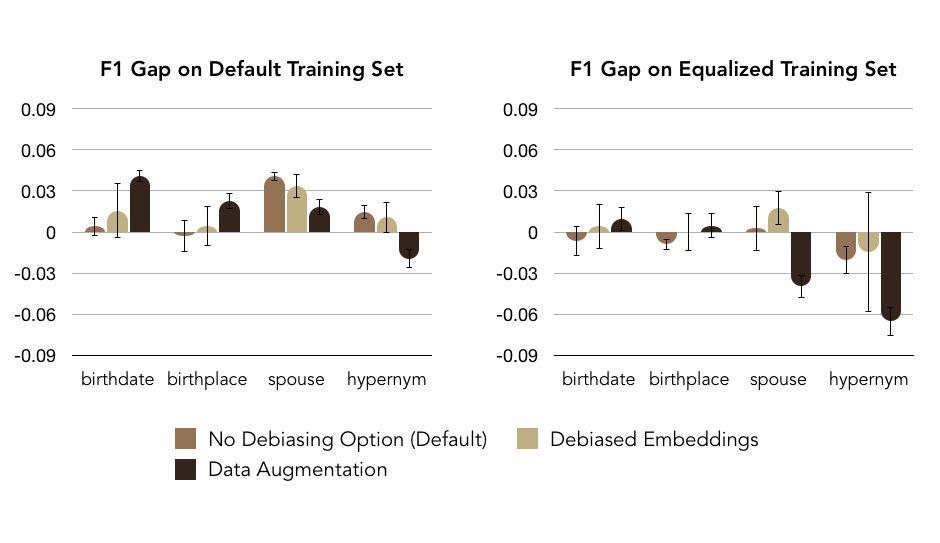}
  \caption{Bias in relation extraction model on each relation as measured by $male - female$ F1 score gender gap (used to calculate disparity score) for the  default training set without modifications (left) and equalized training set (right). This is evaluated on the model with No Debiasing and two bias mitigation methods: debiased embeddings and data augmentation. The experiment is run five times. We give the mean values and standard error bars.}
  \label{disparity_scores}
\end{figure*}


\paragraph{Word Embedding Debiasing} Word embeddings can encode gender biases \cite{bolukbasi2016man,caliskan2017semantics,garg2018word} and this  can affect bias in downstream predictions for models using the embeddings \cite{zhao2018gender, font2019equalizing}. In this work, we apply the Hard-Debiasing technique~\cite{bolukbasi2016man}.
We applied Hard-Debiasing to Word2Vec embeddings \cite{mikolov2013distributed}, which we trained on the sentences in WikiGenderBias. When used in conjunction with data augmentation, the embeddings are re-trained on the union of the two corpora. Below, we give metrics used for measuring model performance and bias in our experiments.

\begin{table*}[!t]
\centering
\small
 \begin{tabular}{ @{\ \ }c @{\ \ }c @{\ \ }c @{\ \ }c | cccc@{}}
 \hline
 \# & Equalization & Debiased Embeddings & Data Aug. & EoO $\downarrow$ & PPS Score $\uparrow$ & Macro $F_1$ Score $\uparrow$ & Disparity Score $\downarrow$ \\
 \hline
1 &   &   &   & .012 & .870 & .886 & .016 \\
\hline 

2 & \checkmark &   &   & -.011 & .851 & .860 & 0.010 \\
3 &   & \checkmark &   & .015 & \textbf{.886} & \textbf{.902} & .016 \\
4 &   &   & \checkmark & .014 & .841 & .866 & .026 \\
\hline 

5 & \checkmark & \checkmark &   & \textbf{.001} & .863 & .872 & \textbf{.009} \\
6 & \checkmark &   & \checkmark & -.024 & .805 & .835 & .030 \\
7 &   & \checkmark & \checkmark & .018 & .868 & .891 & .023 \\
\hline 

8 & \checkmark & \checkmark & \checkmark & .006 & .867 & .877 & .010 \\
\hline

\end{tabular}

\caption{PPS Scores when using debiased embeddings and data augmentation with the unequalized, original dataset. We find that using debiased embeddings alone leads to the best PPS score. Other combinations of debiasing parameters lowers either F1 score, disparity score, or both. We bold the best values, which represent the maximum for PPS score and F1 score and minimum for Disparity Score.}
\label{pps_table}
\end{table*}

\subsection{Effectiveness of Bias Mitigation}

We note that by downsampling the training instances to equalize the number of male and female datapoints, the difference in performance on male versus female sentences decreases to almost 0 for every relation aside from hypernym (see Figure \ref{disparity_scores}, right). Additionally, the drop in aggregate is performance is relatively small (see Macro F1, Table \ref{pps_table}). Given that we down-sampled male instances to create this equalized dataset, training on the equalized data was also more efficient.

We also examined the effect of various debiasing techniques.
Table \ref{pps_table} shows the results. Unfortunately, most of these techniques cause a significant performance drop and none of them is effective in reducing the performance gap between genders.  
Interestingly, debiasing embeddings increased aggregate performance by achieving slightly better F1 performance. As none of these mitigation approaches is effective, their combinations are not effective as well. They either lowering Macro F1 or raising Disparity Score or both. 

We further examine the performance of various bias mitigation techniques evaluated in each relation in Figure \ref{disparity_scores}. NRE relies heavily on surface-level cues such as context, the entities, and their positions. Data augmentation might potentially introduce artifacts and biases, causing the NRE system captures unwanted patterns and spurious statistics between contexts. 



\section{Conclusion}

In our study, we create and publicly release WikiGenderBias: the first dataset aimed at evaluating bias in NRE models. We train NRE models on the WikiGenderBias dataset and test them on gender-separated test sets. We find a difference in F1 scores for the spouse relation between predictions on male sentences and female for the model's predictions. We also examine existing bias mitigation techniques and find that naive data augmentation causes a significant performance drop. 

It is an open and difficult research question to build unbiased neural relation extractors. One possibility is that some bias mitigation methods that add noise to the dataset encourage neural relation extraction models to learn spurious correlations and unwanted biases. We encourage future work to dive deeper into this problem.

While these findings will help future work avoid gender biases, this study is preliminary. We only consider binary gender, but future work should consider non-binary genders. Additionally, future work should further probe the source of gender bias in the model's predictions, perhaps by visualizing attention or looking more closely at the model's outputs.

\section{Acknowledgments}

We thank anonymous reviewers for their helpful feedback. This material is based upon work supported in part by the National Science Foundation under IIS Grant 1927554 and Grant 1821415: Scaling the Early Research Scholars Program.

\bibliography{anthology,acl2020}
\bibliographystyle{acl_natbib}

\appendix



\end{document}